\documentclass[manuscript,screen]{acmart}
\usepackage{amsmath,amsfonts}
 \usepackage{amssymb}
\usepackage{algorithmic}
\usepackage{graphicx}
\usepackage{textcomp}
\usepackage{xcolor}
\usepackage{CJKutf8} 
\usepackage{float}
\usepackage{multicol}
\usepackage{multirow}
\usepackage{makecell}
\usepackage{hyperref}
\newcommand{\blue}[1]{{\color{black}{#1}}}

\AtBeginDocument{%
  }

\setcopyright{acmcopyright}
\copyrightyear{2024}
\acmYear{2024}
\acmVolume{23}
\acmNumber{4}
\acmDOI{10.1145/3652160}

\begin{document}

\title{Medical Question Summarization with Entity-driven Contrastive
Learning}

\author{Wenpeng Lu*, Sibo Wei}

\affiliation{
    \institution{Key Laboratory of Computing Power Network and Information Security, Ministry of Education, Shandong Computer Science Center (National Supercomputer Center in Jinan), Qilu University of Technology (Shandong Academy of Sciences); Shandong Provincial Key Laboratory of Computer Networks, Shandong Fundamental Research Center for Computer Science}
    \city{Jinan}
    \country{China}
}
\email{wenpeng.lu@qlu.edu.cn, sibo.wei@foxmail.com}

\author{Xueping Peng}

\affiliation{
    \institution{Australian Artificial Intelligence Institute, University of Technology Sydney}
    \city{Sydney}
    \country{Australia}
}
\email{xueping.peng@uts.edu.au}

\author{Yi-Fei Wang}
\affiliation{
    \institution{Affiliated Hospital of Shandong University of Traditional Chinese Medicine}
    \city{Jinan}
    \country{China}
}
\email{71000686@sdutcm.edu.cn}

\author{Usman Naseem}
\affiliation{
    \institution{School of Computing, Macquaire University}
    \city{Sydney}
    \country{Australia}
}
\email{usman.naseem@mq.edu.au}

\author{Shoujin Wang}
\affiliation{
    \institution{Data Science Institute, University of Technology Sydney}
    \city{Sydney}
    \country{Australia}
}
\email{shoujin.wang@uts.edu.au}
\authorsaddresses{Authors’ addresses: Wenpeng Lu, Sibo Wei, Key Laboratory of Computing Power Network and Information Security, Ministry of Education, Shandong Computer Science Center (National Supercomputer Center in Jinan), Qilu University of Technology (Shandong Academy of Sciences); Shandong Provincial Key Laboratory of Computer Networks, Shandong Fundamental Research Center for Computer Science, Jinan, China, wenpeng.lu@qlu.edu.cn, sibo.wei@foxmail.com; Xueping Peng, Australian Artificial Intelligence Institute, University of Technology Sydney, Sydney, Australia, xueping.peng@uts.edu.au; Yi-Fei Wang, Affiliated Hospital of Shandong University of Traditional Chinese Medicine, Jinan, China, 71000686@sdutcm.edu.cn; Usman Naseem, School of Computing, Macquaire University, Sydney, Australia, usman.naseem@mq.edu.au; Shoujin Wang, Data Science Institute, University of Technology Sydney, Sydney,
Australia, shoujin.wang@uts.edu.au. *Corresponding author: Wenpeng Lu.}

\renewcommand{\shortauthors}{Lu et al.}
\begin{abstract}
By summarizing longer consumer health questions into shorter and essential ones, medical question-answering systems can more accurately understand consumer intentions and retrieve suitable answers. However, medical question summarization is very challenging due to obvious distinctions in health trouble descriptions from patients and doctors. Although deep learning has been applied to successfully address the medical question summarization (MQS) task, two challenges remain: how to correctly capture question focus to model its semantic intention, and how to obtain reliable datasets to fairly evaluate performance. To address these challenges, this paper proposes a novel medical question summarization framework based on \underline{e}ntity-driven \underline{c}ontrastive \underline{l}earning (ECL). ECL employs medical entities present in frequently asked questions (FAQs) as focuses and devises an effective mechanism to generate hard negative samples. This approach compels models to focus on essential information and consequently generate more accurate question summaries. Furthermore, we have discovered that some MQS datasets, such as the iCliniq dataset with a 33\% duplicate rate, have significant data leakage issues.
To ensure an impartial evaluation of the related methods, this paper carefully examines leaked samples to reorganize more reasonable datasets. Extensive experiments demonstrate that our ECL method outperforms the existing methods and achieves new state-of-the-art performance, i.e., 52.85, 43.16, 41.31, 43.52 in terms of ROUGE-1 metric on MeQSum, CHQ-Summ, iCliniq, HealthCareMagic dataset, respectively. The code and datasets are available at \href{https://github.com/yrbobo/MQS-ECL}{https://github.com/yrbobo/MQS-ECL.}
\end{abstract}



\maketitle

\section{Introduction}
With the rise of online medical services, more consumers are turning to medical websites to seek answers to their health questions, which requires the assistance of automatic medical question-answering systems (MQAs)\citep{zhang2020chinese}. To provide correct answers, MQAs must accurately understand the intent of questions \citep{zhang2022focus}, which is a critical and challenging task. Consumer health questions are typically submitted by patients and often contain extraneous information, such as patient history, making it difficult for MQAs to retrieve relevant answers. Additionally, consumers may use vocabulary distinct from that of medical professionals, further complicating the task of understanding the questions \citep{mrini2021gradually}.
As a result, researchers have proposed various solutions, such as query relaxation~\citep{lei2020expanding}, question entailment~\citep{ben2019question, mrini2021joint}, and question summarization~\citep{mrini2021gradually, zhang2022focus}. Among these, question summarization has gained increasing attention from the research community.

\begin{table}
    \centering
    \caption{An sample for medical question summarization in MeQSum dataset, where question focus are highlighted with red color.}
    \begin{tabular}{|l|}
        \hline
        \parbox{0.95\linewidth}{\textbf{Input question: consumer health question (CHQ):} \\
        SUBJECT: \textcolor{red}{shingles} MESSAGE: I am having symptoms of \textcolor{red}{shingles}, no rash or blisters, is it too late to get the \textcolor{red}{vaccine}? I have had the chicken pox and take acyclovir on a as needed basis for blisters I get on my buttocks.} \\
        \hline
        \parbox{1\linewidth}{\textbf{Reference summary: frequently asked question (FAQ):}\\
        I am having symptoms of \textcolor{red}{shingles}; is it too late to get the \textcolor{red}{vaccine}?}\\
        \hline
        \hline
        \makecell[l]{\textbf{Summary by BART (baseline):} \\
        What are the symptoms of \textcolor{red}{shingles}?
        }\\ \hline
        \makecell[l]{\textbf{Summary by QFCL (baseline):}\\
        What are the treatments for \textcolor{red}{shingles}?}\\ \hline
        \makecell[l]{\textbf{Summary by our model (ECL):}\\
        Is it too late to get the \textcolor{red}{vaccine} for \textcolor{red}{shingles}?
        }\\ \hline
    \end{tabular}
    \label{tab:intro}
\end{table}

Medical question summarization (MQS) is the task of summarizing longer consumer health questions (CHQs) into shorter ones, i.e., frequently asked questions (FAQs), which contain the essential intent of original CHQs. Compared with lengthy CHQs, FAQs focus on the key information of health questions and filter out unnecessary and redundant information, which is helpful for MQAs to retrieve correct answers. An example of MQS is shown in Table \ref{tab:intro}.

In recent years, deep learning has gained exceptional successes in the task of text summarization\citep{katwe2023methodical, k2023abstractive, zhang2023ga, zhao2023softmax, vo2021se4exsum}. For medical question summarization, the existing approaches can be categorized into three groups: Seq2Seq-based methods, RL (reinforcement learning)-enhanced methods, and CL (contrastive learning)-enhanced methods\citep{ma2022multi}. The original Seq2Seq-based methods train the encoder and decoder with an attention mechanism to predict the most probable word that emerges next in the summary text sequence \citep{zhang2020pegasus}. Some variants of Seq2Seq, including T5\citep{raffel2020exploring}, PEGASUS\citep{zhang2020pegasus}, ProphetNet\citep{qi2020prophetnet}, BART \citep{lewis2020bart}, pointer-generator network \citep{abacha2019summarization}, and multi-task learning framework \citep{mrini2021gradually}, have also been applied to the MQS task. However, these models rely on maximum likelihood estimation (MLE), which prioritizes the accuracy of masked token prediction over the preservation of semantic similarity or dissimilarity of questions \citep{zhang2022focus}. To address this limitation, RL-enhanced methods attempt to utilize reinforcement learning with question-aware semantic rewards to enhance the ability of Seq2Seq-based framework on the MQS task. These methods create reasonable rewards, such as question-type identification and question-focus recognition, to ensure the quality of generated question summary \citep{yadav2021reinforcement}. However, RL-enhanced methods still suffer from noise gradient estimation problems, which results in unstable training \citep{zhang2022focus}, and require substantial annotated training data for the devised rewards. To overcome the challenges associated with Seq2Seq and RL-enhanced methods, CL-enhanced methods utilize contrastive learning to automatically generate effective negative samples to enhance semantic representations and improve summarization performance \citep{zhang2022focus}. Nevertheless, constructing high-quality hard negative samples remains a challenge for CL-enhanced methods.

For the medical question summarization task, focus words or phrases in health questions are crucial and essential for representing the core intention of questions. This has been verified by the originators of the task, i.e., Ben Abacha et al. \citep{abacha2019summarization}, who state that question focus is very important, and advise training models to pay more attention to the focus words or phrases in health questions so as to avoid the related errors, such as missing or wrongly identifying question focus. We show an example from MeQSum dataset \citep{abacha2019summarization} in Table \ref{tab:intro}, where the summaries generated by two state-of-the-art (SOTA) baselines and our proposed model are listed, with question focuses highlighted in red. According to the input question (CHQ) and reference summary (FAQ), it is obvious that both \verb|shingles| and \verb|vaccine| are question focuses, which are absolutely necessary for accurately representing the consumer intention. However, \verb|vaccine| is missed by both SOTA baselines, i.e, BART and QFCL \citep{lewis2020bart,zhang2022focus}. Besides, missing question focus further induces the wrong identification of question types. 
At present, QFCL is the best SOTA baseline, which assumes the overlap phrases between CHQ and FAQ as question focus, and utilizes them to construct hard negative samples. Although QFCL can identify question focus via overlap phrases, there are two limitations. One is that as the distinct vocabulary from patients and doctors, the overlap rate is not high (only 68\% on MeQSum dataset). Another is that not all overlap phrases are question focus, which may induce some noise. Accurately capturing question focus and generating hard negative samples are crucial for further improvement on the MQS task. 

As medical question summarization is an emerging task in natural language processing, having enough and reliable data is another important and urgent issue. There are four popular datasets available, including MeQSum \citep{abacha2019summarization}, CHQ-Summ \citep{yadav2021nlm}, iCliniq and HealthCareMagic \citep{mrini2021joint}. Among them, the MeQSum dataset is manually constructed by selecting CHQs together with their summaries from the U.S. National Library of Medicine. The CHQ-Summ dataset is extracted from the healthcare category in Yahoo! Answers L6 corpus. The iCliniq and HealthCareMagic datasets are extracted from MedDialog dataset \citep{mrini2021joint, zeng2020meddialog}. Although the pioneering works strive to construct suitable datasets, according to our empirical investigation, some of them suffer from serious data leakage problems. For example, the duplicate rate of data samples in the iCliniq datasets reaches 33\% (duplicate samples refer to samples with different IDs but identical content), which leads to the overlap of training and test data and makes the evaluation results unreliable. To ensure reliable evaluations of related models on the MQS task, it is crucial to carefully check the datasets and filter out any duplicate samples.

To address the problems mentioned above, we propose a novel framework for medical question summarization called ECL (\underline{e}ntity-driven \underline{c}ontrastive \underline{l}earning). ECL includes two mechanisms to respectively construct hard and simple negative samples/summaries for each training question (CHQ). For hard negative samples, we assume medical entities in reference summaries (FAQs) are question focuses and generate hard negative samples by replacing the original focuses with randomly selected, unrelated entities. For simple negative samples, we randomly select summaries of questions different from the training question. For positive samples, we obtain them with reference summaries (FAQs). With the contrastive learning framework, we bring CHQ/generated summaries and reference summaries (FAQ) closer and push apart CHQ/generated summaries and other negative samples farther apart in semantic representation space. 
With the support of hard negative samples and contrastive learning, ECL can capture question focus and generate question summaries more accurately. 
In addition, we carefully investigated the data samples in the existing MQS datasets, filtered out the duplicated samples, and constructed reliable datasets. This benefits fair comparisons with related works. 

Our contributions are summarized as follows:
\begin{itemize}
    \item We propose ECL (\underline{e}ntity-driven \underline{c}ontrastive \underline{l}earning), a framework for medical question summarization that uses entity-driven contrastive learning. Compared to existing contrastive learning-based methods, ECL employs an easy and effective mechanism to capture the question focus with medical entities and generate hard negative samples more accurately. This is the first work, to our knowledge, that leverages medical entities to identify question focus and enhance performance on MQS tasks.
    
    \item We carefully checked and reorganized the datasets for the MQS task. We filtered out duplicate samples, which effectively avoids the problem of data leakage. With the updated datasets, we evaluate existing baselines and ECL fairly. The revised datasets are reliable and essential for further research in the MQS community.
    
    \item Substantial experiments on four datasets demonstrate that ECL outperforms state-of-the-art methods by capturing question focus and generating medical question summaries more accurately. 
\end{itemize}

\section{Related Work}
\subsection{Medical Question Summarization}
The Medical Question Summarization (MQS) task was introduced by Ben Abacha et al. in 2019 \citep{abacha2019summarization}. The authors manually developed the initial MeQSum dataset for the MQS task and applied Seq2Seq models and a pointer-generator network to produce summaries of CHQ. In 2021, Ben Abacha et al. organized the MEDIQA shared task, which focused on summarization in the medical field. All the submissions employed abstractive summarization with pre-trained models fine-tuning \citep{abacha2021overview}. The difference among the approaches primarily lies in the variety of pre-trained models utilized, including ProphetNet, PEGASUS, and BART \citep{yadav2021nlm, mrini2021ucsd, sanger2021wbi}. Moreover, Mrini et al. proposed to jointly optimize question summarization and entailment tasks with BART by incorporating multi-task learning (MTL) and data augmentation methods to enhance the efficacy of joint optimization \citep{mrini2021joint, mrini2021gradually}. All the aforementioned works were built on Seq2Seq models that were optimized using the Maximum Likelihood Estimation (MLE) approach. However, MLE focuses solely on the accuracy of masked token prediction and does not account for either the semantic similarity or dissimilarity of questions, which adversely affects the quality of the resulting summary \citep{mrini2021gradually}.

To address the limitation, some researchers argue that question focus and type should be noticed, which are valuable for ensuring the generation of semantically valid questions. Yadav et al. proposed to employ reinforcement learning (RL) on MQS with question-aware semantic rewards, which were obtained from two subtasks including question-type identification and question-focus recognition \citep{yadav2021reinforcement}. However, this RL-enhanced method is confused by the problem of noise gradient estimation \citep{mrini2021gradually}. Moreover, it relies on the manually-annotated training data to implement the rewards, which hinders its application on large-scale datasets, such as iCliniq and HealthCareMagic. To further conquer the problem, Zhang et al. proposed the question focus-driven contrastive learning (QFCL) on MQS, which assumed the overlap phrases between CHQ and FAQ as question focuses, and annotated them automatically \citep{zhang2022focus}. According to question focuses, QFCL generated hard negative samples by randomly replacing focus phrases with the other ones. However, QFCL still is confused by two limitations. The overlap rate of phrases in CHQ and FAQ is not high, which means that some questions without overlapped phrases will miss question focuses. Besides, the overlap phrases may be not a true focus, which means that some focuses are annotated wrongly. 

All the above methods are not very satisfactory to solve MQS tasks. Our method ECL assumes medical entities in FAQs are question focuses, and generate hard negative samples by replacing the original focuses with randomly-selected unrelated entities, which can avoid the two limitations of QFCL. 

\subsection{Contrastive Learning}
Contrastive learning (CL) is an effective learning strategy in various domains, such as computer vision and NLP fields \citep{hadsell2006dimensionality, jaiswal2020survey}. Its core idea is to train models to distinguish whether samples are similar or different, which pulls the similar samples closer and pushes the different ones farther apart. CL has been widely applied on a series of NLP tasks, such as text summarization\citep{liu2021simcls,cao2021cliff, liu2022brio, xu2022sequence}, machine translation \citep{yang2019reducing,pan2021contrastive,zhang2022frequency, li2022improving}, question answering \citep{yang2021xmoco, caciularu2022long}, sentence embedding\citep{gao2021simcse,zhang2022mcse, chen2022information,tan2022sentence}, named entity recognition\citep{huang2022copner,das2022container,ying2022label}, and natural language understanding\citep{wang2021cline}. 

To implement CL strategy, most of the former works employ the SimCLR framework proposed by chen et al. \citep{chen2020simple}. SimCLR argued that a huge amount of negative samples is beneficial to improve the effectiveness of CL, while it suffered from the troubles of the heavy computational burden. He et al. presented momentum contrast (MoCo) for unsupervised representation learning \citep{he2020moco}. MoCo views CL as building a dynamic dictionary, where keys in the dictionary are sampled and represented by an encoder. The encoder is trained by performing dictionary look-up, i.e., an encoded query should be similar to its key and different from the others. MoCo maintains the dictionary as a queue, which can store a large number of negative samples. Besides, MoCo proposes a momentum update mechanism to slowly update the key encoder from the query encoder. MoCo can achieve better performance on MQS tasks, which has been verified by QFCL \citep{zhang2022focus}. Therefore, we adopt MoCo as the fundamental architecture to realize our method. 

\section{Methodology}
\begin{figure*}[htbp]
\centering
\includegraphics[width=1\textwidth]{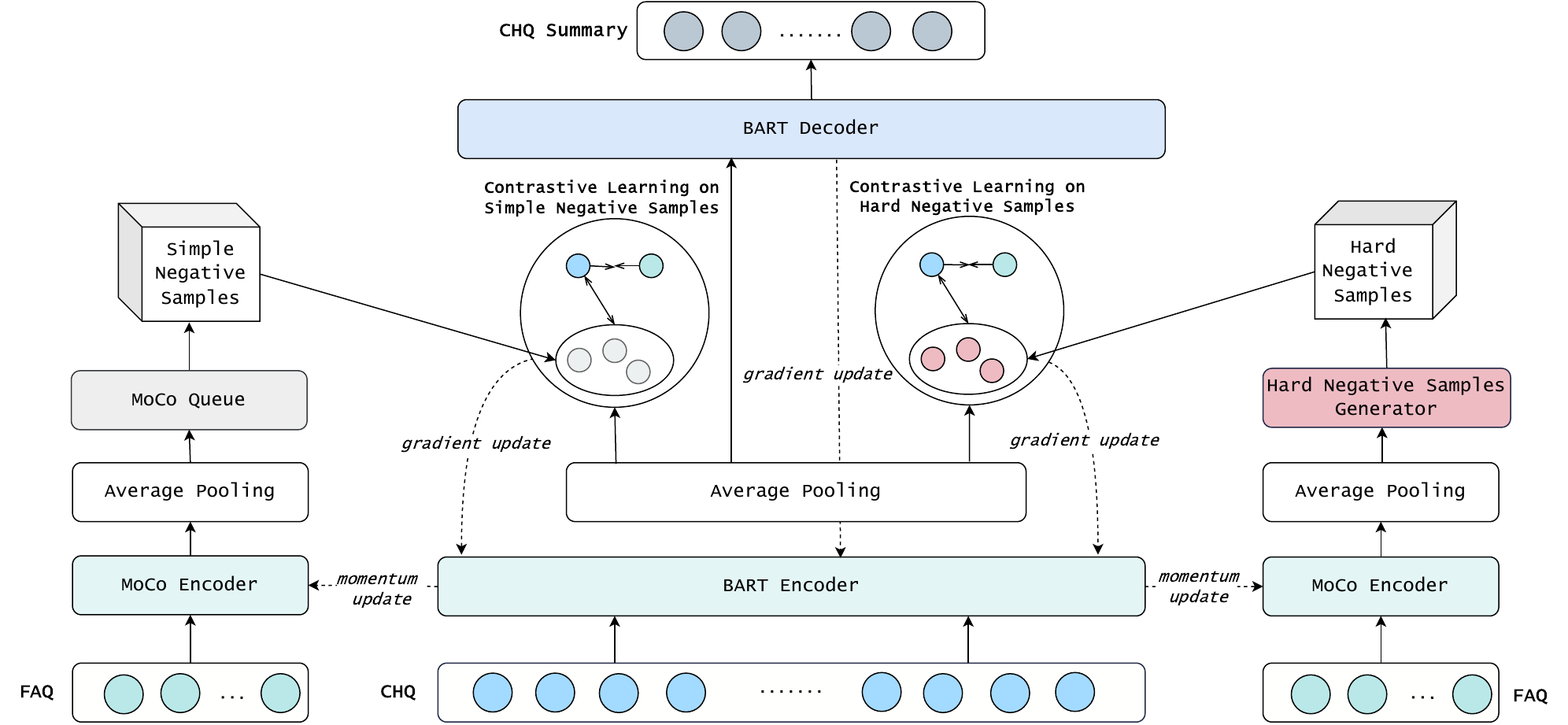}
\caption{Framework of the proposed medical question summarization with entity-driven contrastive learning (ECL). It involves basic summarization module (BART), contrastive learning module (MoCo), simple negative sample module and hard negative sample module.}
\label{fig:model}
\end{figure*}

In this section, we first present the task definition of MQS, then introduce the overall framework of our proposed ECL model, followed by the details of each core functional module of ECL.

\subsection{Task Definition}
The medical question summarization (MQS) task is to summarize a longer consumer health question (CHQ) into a shorter one, i.e., a frequently asked question (FAQ), which contains the essential intention of the original CHQ. Specifically, given a lengthy question $C$, our goal is to learn a function $f$  to generate its summary $S$, as below:
\begin{equation}
S\leftarrow f\left( C \right), 
\label{eq:taskdefination}
\end{equation}
where $C$ is a CHQ asked by a consumer and may contain much redundant information, $S$ is a FAQ containing the intrinsical intention of the CHQ with as shorter as words. Compared with $C$, $S$ is more beneficial for question answering system to retrieve correct answers for the consumer.

\subsection{Framework of our ECL model}
The framework of our proposed ECL model is shown in Fig. \ref{fig:model}, which consists of four core functional modules, including a basic summarization module, contrastive learning module, simple negative sample module, and hard negative sample module. For the basic summarization module, we adopt pre-trained BART \citep{lewis2020bart} as the basic summarization model, which employs MLE and cross-entropy to generate question summaries. For the contrastive learning module, we adopt MoCo \citep{he2020moco} as the fundamental architecture, which employs a queue to store a large number of negative samples, and utilizes a momentum update mechanism to update the key encoder from the query encoder. For a simple negative sample module, for a CHQ $C_i$, we randomly select multiple summaries unmatched with $C_i$ as its simple negative samples. For the hard negative sample module, we assume medical entities in FAQs are question focuses and generate hard negative samples by replacing the original focuses with randomly-selected unrelated entities. Finally, the loss of cross-entropy and contrastive learning based on simple/hard negative samples are combined together to jointly optimize our ECL model.

\subsection{Basic Summarization Module}
In essence, MQS is a subtask of text generation, which has recently been dominated by BART. BART \citep{lewis2020bart} is a transformer-based sequence-to-sequence model, which has been shown to be particularly effective when fine-tuned for text generation tasks. Naturally, BART also demonstrates its superiority on MQS task \citep{mrini2021joint,zhang2022focus,mrini2021gradually}. Following the existing works, we adopt pre-trained BART as our basic summarization model. 

For BART, its standard training algorithm is maximum likelihood estimation (MLE). For the specific $i$-th training sample $\{ {C_i},S_i^*\}$, MLE is equivalent to minimizing the sum of the negative likelihood of the $l$ tokens $\{ s_1^*, \cdots,s_j^*,\cdots,s_l^*\} $ in the reference summary ${{S}^{*}}$, i.e., to optimize the cross-entropy loss:
\begin{equation}
{{\mathcal{L}}_{ce}} =  - \sum\limits_{j = 1}^l {\sum\limits_s^{} {{p}(s|C,S_{ < j}^*)\log {p_{{f_\theta }}}(s|C,S_{ < j}^*;\theta )} },
\label{eq:crossentropy}
\end{equation}
where $S_{ < j}^*$ refers to the partial reference sequence $\{ s_0^*, \cdots,s_{j - 1}^*\} $ and $s_0^*$ is a pre-defined start token. $\theta$ refers to the parameters of $f$ and ${{p}_{{{f}_{\theta }}}}$ is the probability distribution entailed by these parameters. \blue{$p$ is the one-hot distribution optimized using label smoothing: }
\blue{\begin{equation}
    p(s|C,S_{ < j}^*) = \left\{ \begin{gathered}
  1 - \lambda ,s = s_j^* \hfill \\
  \frac{\lambda }{{\alpha  - 1}},s \ne s_j^* \hfill \\ 
\end{gathered}  \right.,
\end{equation}}
\blue{where $\lambda$ is the probability mass and $\alpha$ is the size of the dictionary \citep{liu2022brio}}.

\subsection{Contrastive Learning Module}

In order to implement contrastive learning, we adopt MoCo as the fundamental architecture \citep{he2020moco}. MoCo views CL as a dictionary look-up and proposes momentum contrast for unsupervised representation learning. As its powerful ability to support a large size dictionary, MoCo can handle a huge amount of negative samples, which makes it achieve great success \citep{zhang2022focus}. Therefore, we choose MoCo as the basic CL architecture to handle the negative samples. MoCo implements momentum contrast as below:

\subsubsection{Dictionary as a queue} MoCo maintains the dictionary as a queue of samples, whose length is flexible and not limited by the size of the mini-batch. The dictionary is dynamic, which means that the keys are randomly sampled and the key encoder evolves during training. The samples in the dictionary are replaced progressively, where the current mini-batch is enqueued while the oldest mini-batch is dequeued. With the support of the queue, we can use a large number of simple negative samples. Note that hard negative samples will not enter the queue.

\subsubsection{Momentum update} The utilization of the large queue results in the impossibility of updating the key encoder with backpropagation. MoCo devises a momentum update mechanism to solve the problem. Following the work of QFCL \citep{zhang2022focus}, we make the key encoder ${E_k}$ in MoCo adopt the same structure as the BART encoder ${E_q}$. The parameters of ${E_k}$ updated slowly by ${E_q}$:
\begin{equation}
    {\theta _k} \leftarrow m{\theta _k} + (1 - m){\theta _q},
    \label{eq:momentum}
\end{equation}
where $m \in [0,1)$ is a momentum coefficient.

\subsection{Simple Negative Sample Module}
\label{sec:scl}

For a specific training sample $\{ {C_i},S_i^*\} $, ${C_i}$ (CHQ) should be semantically similar to its reference summary $S_i^*$ (FAQ) and different from the summary $S_j^*$ of another question \citep{zhang2022focus}. Therefore, we regard $S_i^*$ as the positive sample of ${C_i}$ and randomly select $S_j^*$ from other training samples as the simple negative samples of ${C_i}$, as shown in the right panel of Fig. \ref{fig:negativesamples}. 

The objective function of simple contrastive learning is defined below:
\begin{equation}
    \blue{{\mathcal{L}_{scl}} =  - \log \frac{{{e^{sim({\mathbf{R}_{{C_i}}},{\mathbf{R}_{S_i^*}})/\tau }}}}{{\sum\nolimits_{{\mathbf{R}_{S_j^*}} \in \mathbb{Q}} {{e^{sim({\mathbf{R}_{{C_i}}},{\mathbf{R}_{S_j^*}})/\tau }}} }},}
    \label{eq:loss-simplenegativesamples}
\end{equation}
\blue{where ${{\mathbf{R}_{{C_i}}}}$ refers to the sentence representation of the $i$-th CHQ obtained from BART encoder ${E_q}$, ${{\mathbf{R}_{S_i^*}}}$ and ${{\mathbf{R}_{S_j^*}}}$ are obtained from MoCo key encoder ${E_k}$. ${sim({\mathbf{R}_{{C_i}}},{\mathbf{R}_{S_i^*}})}$ is the cosine similarity $\frac{{\mathbf{R}_{{C_i}}^{\top}{\mathbf{R}_{S_i^*}}}}{{||{\mathbf{R}_{{C_i}}}|| \cdot ||{\mathbf{R}_{S_i^*}}||}}$ and $\tau$ is a temperature hyper-parameter. $\mathbb{Q}$ is the queue that stores simple negative samples, whose size is a hyper-parameter $l_q$.}

\subsection{Hard Negative Sample Module}
\label{sec:hcl}
\begin{figure*}[htbp]
\centering
\includegraphics[width=1\textwidth]{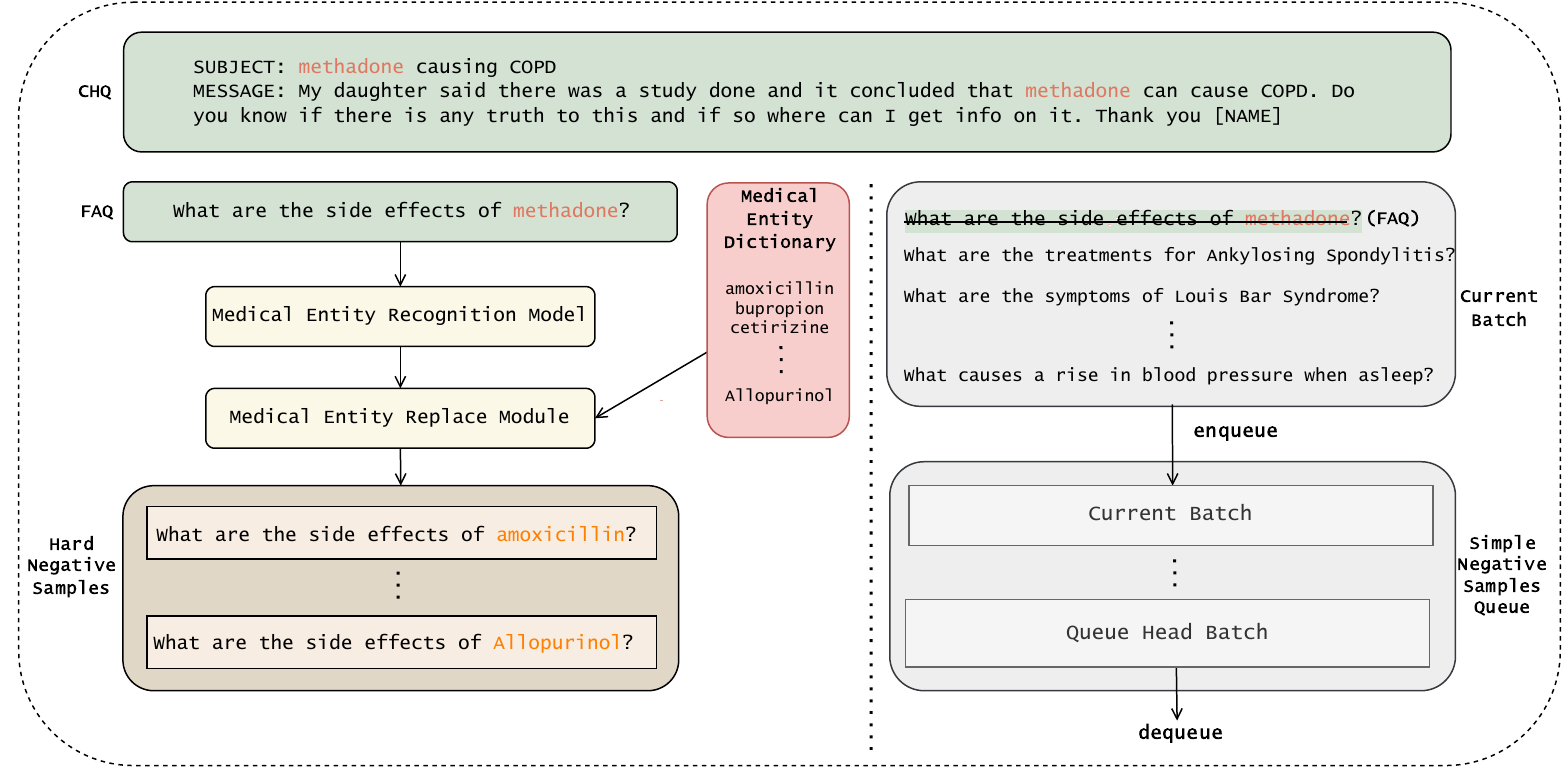}
\caption{Generation of hard and simple negative samples.}
\label{fig:negativesamples}
\end{figure*}

As claimed by the originators of MQS task \citep{abacha2019summarization}, how alleviating the missing problem of question focus is critical for generating a satisfied summary for CHQ. According to question focus, to construct high-quality hard negative samples is an effective way to solve the problem. As the best baseline in MQS, QFCL employs contrastive learning to improve the quality of question summary, which assumes the overlap phrases between CHQ and FAQ as question focuses, and constructs hard negative samples \citep{zhang2022focus}. Different from QFCL, we assume medical entities in FAQ as question focuses. The generation method of hard negative samples is shown in the left panel of Fig. \ref{fig:negativesamples}.

\subsubsection{Question Focus Identification} 
Imagine a scenario in which a patient describes his trouble to a doctor. Although the description of the patient (CHQ) may be lengthy and contain abundant peripheral information, the experienced doctor is able to summarize the patient's question (CHQ) into the reference question (FAQ) according to one or several key medical entities. Intuitively, medical entities can be viewed as question focuses.

In order to automatically recognize medical entities from the medical question, we adopt the \verb|Stanza| toolkit \citep{qi2020stanza}, which supports biomedical and clinical named entity recognition \footnote{https://stanfordnlp.github.io/stanza/}. For each reference summary in the training dataset, we annotate medical entities with \verb|Stanza|, which are labeled as the question focuses. 

\blue{In addition, to facilitate the generation of hard negative samples, we traverse the entire training dataset to collect all medical entities to build a medical entity dictionary, marked as $\mathbb{D}$.}

\subsubsection{Hard Negative Sample Generation}
\blue{We generate hard negative samples by replacing medical entities in a reference summary with unrelated entities. Specifically, for a reference summary of CHQ, marked as ${S_i^*}$, we first collect the medical entities contained in ${S_i^*}$, marked as $\mathbb{M} = \{ {M_1}, \cdots ,{M_t}\} $. Then, we remove the entities in $\mathbb{M}$ from the medical entity dictionary $\mathbb{D}$ to obtain ${\mathbb{D}'}$. For the reference summary ${S_i^*}$, we replace each medical entity with one entity randomly selected from ${\mathbb{D}'}$ to generate one hard negative sample.} To suppose that the number/hyper-parameter of hard negative samples of each question focus is set to $t$, we repeat the former replacement operation $x$ times, to obtain $t \times x $ hard negative samples for ${S_i^*}$. An example of hard negative sample generation is shown in the left panel of Fig. \ref{fig:negativesamples}.

The loss function of contrastive learning on hard negative samples is defined below:
\begin{equation}
    \blue{{{\mathcal{L}}_{hcl}} =  - \log \frac{{{e^{sim({\mathbf{R}_{{C_i}}},{\mathbf{R}_{S_i^*}})/\tau }}}}{{\sum\nolimits_{{\mathbf{R}_h} \in \mathbb{H}} {{e^{sim({\mathbf{R}_{{C_i}}},{\mathbf{R}_h})/\tau }}} }},}
    \label{eq:loss-hardnegative}
\end{equation}
\blue{where ${\mathbf{R}_h}$ refers to the sentence representation of hard negative sample $h$ obtained from MoCo key encoder ${E_k}$, $\mathbb{H}$ denotes the set containing hard negative samples, whose size is a hyper-parameter $n_h$.}

With the support of the hard negative samples, our proposed ECL model can pay more attention to the question focuses (medical entities), and alleviates the problem of missing question focus or incorrect identification.

\subsection{Objective Function}
We combine the loss of cross-entropy and contrastive learning together as the final objective function, described as: 
\begin{equation}
    {\mathcal{L} } = {{\mathcal{L} }_{ce}} + {{\mathcal{L}}_{scl}} + {{\mathcal{L}}_{hcl}},
\end{equation}
where ${\mathcal{L}}_{ce}$ is from Eq. \eqref{eq:crossentropy}, which ensures the generation ability of the pretrained BART. ${{\mathcal{L}}_{scl}}$ and ${{\mathcal{L}}_{hcl}}$ are from Eq. \eqref{eq:loss-simplenegativesamples} and Eq. \eqref{eq:loss-hardnegative}, which help ECL to enhance the quality of generated question summaries.

\section{Experiment Results and Discussion}

In this section, extensive experiments on four public datasets are carried out to evaluate the performance of the proposed ECL model, which is compared with nine popular state-of-the-art baselines. In addition, the effectiveness of key modules is investigated by an ablation study. Lastly, the identification rate and accuracy of question focus are analyzed, followed by a case study and discussion. 

\subsection{Datasets}
We conduct the experiments on four medical question summarization datasets, i.e., MeQSum\footnote{https://github.com/abachaa/MeQSum}, CHQ-Summ\footnote{https://github.com/shwetanlp/Yahoo-CHQ-Summ}, iCliniq and HealthCareMagic\footnote{https://github.com/UCSD-AI4H/Medical-Dialogue-System}.

\begin{itemize}
    \item \textbf{MeQSum}. It is the first dataset that is created by the originators of MQS task, i.e., Ben Abacha et al. \citep{abacha2019summarization}. The questions are selected from a collection distributed by the U.S. National Library of Medicine \citep{kilicoglu2018semantic}. The dataset contains 1,000 patient health questions (CHQ). Each question is annotated with a question summary by medical experts. 
    \item \textbf{CHQ-Summ}. It is created by Yadav et al. \citep{yadav2022chq}. It contains 1,507 CHQs together with their summaries, which are extracted from the healthcare category in the Yahoo! Answers L6 corpus\footnote{https://webscope.sandbox.yahoo.com/catalog.php?datatype=l\&did=11}, and are manually annotated by six experts in medical informatics and medicine. 
    \item \textbf{iCliniq}. It is created by Mrini et al. \citep{mrini2021joint}, extracted from MedDialog \citep{zeng2020meddialog}. MedDialog data is crawled from multiple sources, including \textit{iclinic.com} and \textit{healthcaremagic.com}, which are two online platforms of healthcare services. Here, the iCliniq dataset is from the former platform, which contains 31,064 samples. The summaries in this dataset are written by patients. However, in our experiments, we find that this dataset suffers from a serious data leakage problem. We check the data sample carefully and verify that there exist 10,368 \blue{duplicate} samples. In order to make the dataset more reliable, we remove the \blue{duplicate} ones and obtain an updated dataset including 20,696 samples. 
    \item \textbf{HealthCareMagic}. It also is created by Mrini et al. \citep{mrini2021joint}, extracted from MedDialog \citep{zeng2020meddialog}. The main differences with the former iCliniq lie in two aspects. One is that this dataset is from the second online platform, i.e., \textit{healthcaremagic.com}. Another is that the summaries in this dataset are written in a formal style, instead of the patient-written summaries in iCliniq. The original HealthCareMagic dataset contains 226,395 samples. For sake of fairness, we also check the dataset and find 523 \blue{duplicate} items. We remove the \blue{duplicate} ones and obtain an updated dataset including 225,872 samples. 
\end{itemize}

\blue{In order to illustrate the issue of data leakage problem in iCliniq and HealthCareMagic datasets, we show a pair of duplicate samples from iCliniq in Table \ref{tab:iCliniq_sample}. It is obvious that though the two samples have different IDs, their contents including CHQ and FAQ are identical, which are duplicate samples. We have removed such samples from both iCliniq and HealthCareMagic datasets. In detail, for the iCliniq/HealthCareMagic dataset, the size of original dataset is 31,064/226,395 with 10,368/523 duplicate samples; once the duplicate ones are removed, the size of revised dataset changes to 20,696/225,872. 
}

\begin{table*}[h]
    \centering
    \resizebox{\linewidth}{!}{
        \blue{\begin{tabular}{|c|}
    \hline
    \makecell[l]{\textbf{ID}: 0 \\
    \textbf{CHQ}: Hi doctor, I am just wondering what is abutting and abutment of the nerve root means in a back issue. Please \\ explain. What treatment is required for annular bulging and tear? \\
    \textbf{FAQ}: What does abutment of the nerve root mean?}
    \\
    \hline
    \makecell[l]{\textbf{ID}: 15 \\
    \textbf{CHQ}: Hi doctor, I am just wondering what is abutting and abutment of the nerve root means in a back issue. Please \\ explain. What treatment is required for annular bulging and tear? \\
    \textbf{FAQ}: What does abutment of the nerve root mean?
     } \\
    \hline
\end{tabular}}
    }
    \caption{\blue{Duplicate Samples in the iCliniq Dataset.}}
    \label{tab:iCliniq_sample}
\end{table*}

Following the settings in existing work \citep{zhang2022focus, mrini2021gradually}, we split these datasets and show the statistics in Table ~\ref{tab:datasets}. \blue{For the MeQSum and CHQ-Summ datasets, we maintained their original split ratios. As for the iCliniq and HealthCareMagic datasets, we removed duplicates, shuffled the entire dataset, and then split it in an 8:1:1 ratio.}
\begin{table}[htbp]
\caption{Statistics of Datasets.}
\begin{center}
\begin{tabular}{lrrr|rr}
        \hline
         \multirow{2}{*}{Datasets} & \multicolumn{3}{c|}{Examples} & \multicolumn{2}{c}{Avg. Words} \\
         \cline{2-6}
         & Train & Valid & Test & CHQ & FAQ \\ \hline
         MeQSum &400 &100 &500 &70 &12 \\
         CHQ-Summ &800 &300 &407 &176 &13 \\
         iCliniq &16,556 &2,069 &2,071 &114 &13 \\
         HealthCareMagic & 180,697 & 22,587 & 22,588 & 93&11 \\
         \hline
    \end{tabular}
\label{tab:datasets}
\end{center}
\end{table}

\subsection{Baselines}
\label{sec:baselines}
We choose some recent competitive state-of-the-art methods as our baselines.

\begin{itemize}
    \item  {\textbf{T5} \citep{raffel2020exploring}. Text-to-Text-Transfer-Transformer (T5) model proposes reframing all NLP tasks into a unified text-to-text-format where the input and output are always text strings. This formatting makes T5 model fit for multiple NLP tasks, including abstractive summarization.}
    \item {\textbf{PEGASUS} \citep{zhang2020pegasus}. PEGASUS proposes a transformer-based model for abstractive summarization. It utilizes a special self-supervised pre-training objective called gap-sentences generation (GSG), which is designed to perform well on summarization-related downstream tasks. }
    \item {\textbf{ProphetNet} \citep{qi2020prophetnet}. ProphetNet is a sequence-to-sequence pre-training model that introduces a novel self-supervised objective named future $n$-gram prediction and the proposed $n$-stream self-attention mechanism.}
    \item \textbf{PG + Data Augmentation} \citep{abacha2019summarization}. It explores data augmentation methods to augment training set incrementally, and employs the pointer-generator (PG) \citep{see2017get} network to generate the summary of CHQ.
    \item \textbf{ProphetNet + QTR + QFR} \citep{yadav2021reinforcement}. It adopts ProphetNet\citep{qi2020prophetnet} and reinforcement learning framework to solve MQS task, which devises two question-aware semantic rewards: question-type identification reward (QTR) and question-focus recognition reward (QFR).
    \item \textbf{Data-Augmented Joint Learning} \citep{mrini2021joint}. It proposes a data augmentation mechanism to use one single dataset to jointly train two tasks, i.e., question summarization and recognizing question entailment. 
    \item \textbf{RQE + MTL + Data Augmentation} \citep{mrini2021gradually}. It verifies the equivalence between the tasks of medical question summarization and recognizing question entailment (RQE), and proposes a multi-task learning (MTL) method with data augmentation for medical question understanding. 
    \item \textbf{BART} \citep{lewis2020bart}. It is a transformer-based sequence-to-sequence pretrained model with a bidirectional encoder and an autoregressive decoder. It has been verified to be particularly effective when fine tuned for text generation task, which has been widely adopted as the basic architecture by MQS models.
    \item \textbf{QFCL} \citep{zhang2022focus}. It is the best and state-of-the-art work on MQS task, which proposes a question focus-driven contrastive learning framework (QFCL) with a two-anchor strategy and a novel method of generating hard negative samples.
    
\end{itemize}

\subsection{Parameter Settings and Evaluation Metrics}
We utilize BART-large \citep{lewis2020bart} in HuggingFace\footnote{https://huggingface.co/facebook/bart-large} as our pretrained model. The learning rate is set to 1e-5 the same as QFCL \citep{zhang2022focus}. The batch size is set to 16. The parameters of Adam optimizer, i.e., $\beta_1$ and $\beta_2$, are set to 0.9 and 0.999. The number of hard negative samples $n_h$ is set to 128. In hard negative sample generation, the sample size $X$ is set to 128 for MeQSum and CHQ-Summ datasets, 256 for the iCliniq dataset, and 512 for the HealthCareMagic dataset. For MoCo \citep{he2020moco}, the momentum coefficient $m$ is 0.999, the temperature $\tau$ is 0.07 and the queue size $l_q$ is 4096. \blue{Under the aforementioned settings, our model requires approximately 12.5GB of GPU memory during training.} Our experiments were run with one NVIDIA A100 40GB GPU. 

Following previous work, we adopt ROUGE \citep{lin2004rouge} as the evaluation metric. R1, R2, and RL metrics denote the F$_1$-scores of ROUGH-1, ROUGH-2, and ROUGH-L, respectively.

\subsection{Experimental Results}
\label{sec:experimentalresult}
\begin{table*}[h]
\caption{Experimental results on four MQS datasets. The best performance
is boldfaced, and the suboptimal one is underlined. \\$^*$ refers to the
improvement made by ECL over the best-performing baseline.}
\centering
\resizebox{\linewidth}{!}{
\begin{tabular}{l|c|c|c|c|c|c|c|c|c|c|c|c}
\hline
\multirow{2}{*}{Model} &
\multicolumn{3}{c|}{MeQSum} & 
\multicolumn{3}{c|}{CHQ-Summ}  &
\multicolumn{3}{c|}{iCliniq} & 
\multicolumn{3}{c}{HealthCareMagic} 
\\ \cline{2-13}
 & 
\multicolumn{1}{c|}{R1} & 
\multicolumn{1}{c|}{R2} & 
\multicolumn{1}{c|}{RL} & 
\multicolumn{1}{c|}{R1} & 
\multicolumn{1}{c|}{R2} & 
\multicolumn{1}{c|}{RL} & 
\multicolumn{1}{c|}{R1} & 
\multicolumn{1}{c|}{R2} & 
\multicolumn{1}{c|}{RL} & 
\multicolumn{1}{c|}{R1} & 
\multicolumn{1}{c|}{R2} & 
\multicolumn{1}{c}{RL} \\
\hline\hline
T5 & 34.47 & 17.18 & 30.54 & 35.17 & 18.87 & 32.33 & 39.25 & 20.84 & 34.92 & 38.09 & 18.41 & 34.99 \\ \hline
PEGASUS & 43.18 & 26.15 & 40.87 & 35.89 & 18.86 & 33.27 & 36.09 & 18.30 & 31.45 & 35.17 & 15.75 & 30.15 \\ \hline
ProphetNet & 44.40 & 26.93 & 41.57 & 40.46 & 22.80 & 38.13 & 39.13 & 20.15 & 34.01 & 30.34 & 12.00 & 26.00 \\ \hline
PG + Data Augmentation$^\diamondsuit$   & 44.16 & 27.64 & 42.78     & -  &  - &  -   & - & -  & -   & -  & - &  -               \\ 
\hline
ProphetNet + QTR + QFR$^\diamondsuit$   & 45.52 & 27.54 & 48.19     & -  &  - &  -   & - & -  & -   & -  & - & -                \\
\hline
Data-Augmented Joint Learning$^\diamondsuit$ &  48.50& 29.70& 44.90  & -   & -  & -  &  -   &   - &  -   &    -  &  -  & -             \\
\hline
RQE + MTL + Data Augmentation$^\diamondsuit$   &  49.20  & 29.50  & 44.80  & -   &-    &-    &-   &-       &-                        &  -   &  -   &  -   \\
\hline
BART$^\diamondsuit$ & 46.17 & 28.05 & 43.75 & 41.96 & 23.45 & 39.62 & 40.67 & 21.83 & 36.20 & 43.08 & 23.38 &  40.27    \\
\hline
QFCL & \underline{51.48}  & \underline{34.16}  & \underline{49.08} & \underline{42.18} &  \underline{23.48}   &   \underline{39.81}  & \underline{40.93}  &  \underline{22.07}   & \underline{36.27}   &   \underline{43.36}  & \underline{23.39}  &  \underline{40.44}                         \\

\hline
ECL (ours)  &  \textbf{52.85}  & \textbf{36.06}  & \textbf{50.48} & \textbf{43.16}  &  \textbf{24.26} &  \textbf{40.46}   & \textbf{41.31}  & \textbf{22.27}  & \textbf{36.68}  & \textbf{43.52}    &   \textbf{23.75}  & \textbf{40.56}   \\
\hline
\hline
Improvement$^*$ &  1.37  & 1.9  & 1.4 & 0.98  &  0.78 &  0.65   & 0.38  & 0.20  & 0.41  & 0.16   &   0.36  & 0.12   \\
\hline
\multicolumn{13}{l}{
\parbox{17.6cm}{$\diamondsuit$ denotes that the experimental results of MeQSum dataset are taken from the original paper, and the other experimental results are all obtained by our own experiments. \textbf{QFCL} is the most state-of-the-art model. Some results on the latter three datasets are missing. The reasons are as follows. \textbf{PG + Data Augmentation} does not disclose their train code. \textbf{ProphetNet + QTR + QFR} requires manual labeling of the dataset, which is infeasible for large size datasets, such as HealthCareMagic. For \textbf{Data-Augmented Joint Learning} and \textbf{RQE + MTL + Data Augmenta}-\textbf{tion}, they do not release the initial form of the raw data and we could not run their train code successfully. Therefore, we do not report their experimental results on CHQ-Summ, iCliniq and HealthCareMagic datasets.}
}
\end{tabular}
}
\label{tab:overallresults}
\end{table*}

{The performance of our proposed ECL model and nine competing models on four public MQS datasets are shown in Table ~\ref{tab:overallresults}. The results marked with $\diamondsuit$ denotes that the experimental results of MeQSum dataset are taken from the original paper, and the other experimental results are all obtained by our own experiments \citep{abacha2019summarization,yadav2021reinforcement,mrini2021joint, mrini2021gradually, zhang2022focus}. For CHQ-Summ, it is a new dataset created in 2022, we re-run all available baselines on it. For iCliniq and HealthCareMagic, we revise them to avoid the problem of data leakage, and we also re-run all available baselines on them to obtain fair results. According to Table
~\ref{tab:overallresults}, we have the following observations. }

{First, T5, PEGASUS, ProphetNet, {ProphetNet + QTR + QFR} and {PG + Data Augmentation} are significantly weaker than the other competing methods based on BART. They are respectively based on pre-trained language models other than BART or pointer-generator network. Although {ProphetNet + QTR + QFR} and {PG + Data Augmentation} have attempted to utilize reinforcement learning  or pointer-generator network to improve the performance, they still be beaten by the other BART-based ones. This demonstrates that the architecture of BART is effective and powerful, which is essential for MQS task. }

Second, {Data-Augmented Joint Learning}, {RQE + MTL + Data Augmentation}, and {BART} are superior to the former {PG + Data Augmentation} and {ProphetNet + QTR + QFR}, while they are inferior to the latter {QFCL}. As the three methods adopt the architecture of BART, leveraging the superiority of BART, they can beat the former two method easily. Although they have attempted to augment the training data, they neglect to construct the hard negative samples with question focuses. This may be the reason that they are beaten by {QFCL}.

Third, QFCL and our proposed ECL model surpass the competing methods significantly. Both methods are built on BART, and utilize the contrastive learning framework. Moreover, both of them implement some strategies to generate high-quality negative samples with question focus, which may be the most important reason for them to achieve better performance. 

Last, our proposed ECL model beats all competing methods and achieves the best performance across all datasets in terms of all metrics. ECL assumes
medical entities in FAQs are question focuses and generates hard negative samples by replacing the original focuses with randomly-selected unrelated entities. Compared with the state-of-the-art QFCL method, ECL can generate more harder negative samples, which are beneficial for model training. This may be the reason for the significant superiority of ECL.

\subsection{Ablation Study}
\label{sec:ablationstudy}

We perform an ablation study to show the effectiveness of different components equipped in ECL. We construct three ablation models for ablation experiments, i.e., ECL$^{-s}$, ECL$^{-h}$ and ECL$^{-s,-h}$. The results are reported in Table ~\ref{tab5}. According to the table, we have the following observations.

\begin{table*}[t]
\caption{Experimental Results of Ablation Study.}
\centering
\resizebox{1\linewidth}{!}{
\begin{tabular}{l|l|l|l|l|l|l|l|l|l|l|l|l}
\hline
\multirow{2}{*}{Model} &
\multicolumn{3}{c|}{MeqSum} & 
\multicolumn{3}{c|}{CHQ-Summ}  &
\multicolumn{3}{c|}{iCliniq} & 
\multicolumn{3}{c}{HealthCareMagic} 
\\ \cline{2-13}
 & 
\multicolumn{1}{c|}{R1} & 
\multicolumn{1}{c|}{R2} & 
\multicolumn{1}{c|}{RL} & 
\multicolumn{1}{c|}{R1} & 
\multicolumn{1}{c|}{R2} & 
\multicolumn{1}{c|}{RL} & 
\multicolumn{1}{c|}{R1} & 
\multicolumn{1}{c|}{R2} & 
\multicolumn{1}{c|}{RL} & 
\multicolumn{1}{c|}{R1} & 
\multicolumn{1}{c|}{R2} & 
\multicolumn{1}{c}{RL} \\
\hline
\hline
ECL &  \makecell[c]{\textbf{52.85}}  & \makecell[c]{\textbf{36.06}}  & \makecell[c]{\textbf{50.48}}  & \makecell[c]{\textbf{43.16}}  &  \makecell[c]{\textbf{24.26}} &  \makecell[c]{\textbf{40.46}}  &  \makecell[c]{\textbf{41.31}} & \makecell[c]{\textbf{22.27}}  & \makecell[c]{\textbf{36.68}}  & \makecell[c]{\textbf{43.52}}   &  \makecell[c]{\textbf{23.75}}  & \makecell[c]{\textbf{40.56}}   \\
\hline

ECL$^{-s}$   & \makecell[c]{52.10\\(-0.75)} & \makecell[c]{34.62\\(-1.44)} &  \makecell[c]{49.42\\(-1.06)}    &  \makecell[c]{42.48\\(-0.68)} &\makecell[c]{24.01\\(-0.25)}  &  \makecell[c]{39.86\\(-0.60)}  & \makecell[c]{40.98\\(-0.33)} & \makecell[c]{22.23\\(-0.04)}  & \makecell[c]{36.47\\(-0.21)}   & \makecell[c]{43.48\\(-0.04)}  & \makecell[c]{23.32\\(-0.43)}  & \makecell[c]{40.37\\(-0.19)}          \\ 
\hline

ECL$^{-h}$   & \makecell[c]{51.72\\(-1.13)} & \makecell[c]{33.91\\(-2.15)} &  \makecell[c]{49.09\\(-1.39)}    & \makecell[c]{42.21\\(-0.95)}  &  \makecell[c]{23.79\\(-0.47)} &  \makecell[c]{39.68\\(-0.78)}   &  \makecell[c]{41.01\\(-0.30)}& \makecell[c]{21.91\\(-0.36)}  & \makecell[c]{36.35\\(-0.33)}   &   \makecell[c]{43.42\\(-0.10)} & \makecell[c]{23.31\\(-0.44)} & \makecell[c]{40.32\\(-0.24)}          \\
\hline
ECL$^{-s,-h}$   & \makecell[c]{46.17\\(-6.68))} & \makecell[c]{28.05\\(-8.01)} & \makecell[c]{43.75\\(-6.73)}     & \makecell[c]{41.96\\(-1.20)}  & \makecell[c]{23.45\\(-0.81)}  &  \makecell[c]{39.62\\(-0.84)}  & \makecell[c]{40.67\\(-0.64)} & \makecell[c]{21.83\\(-0.44)} & \makecell[c]{36.20\\(-0.48)}    & \makecell[c]{43.08\\(-0.44)}  & \makecell[c]{23.38\\(-0.37)} &  \makecell[c]{40.27\\(-0.29)}              \\
\hline

\multicolumn{13}{l}{
\parbox{1\linewidth}{*\textit{-s} means to remove the contrastive learning on simple negative samples, \textit{-h} means to remove the contrastive learning on hard negative samples. The data in brackets means the decrease of the corresponding ablation model, compared with the standard ECL model.}
}
\end{tabular}
}
\label{tab5}
\end{table*} 

First, ECL$^{-s}$ and ECL$^{-h}$ is worse than the standard ECL. This demonstrate that both simple negative samples and hard ones are crucial for our ECL model. Without the negative samples, the performance of ECL will drop significantly.

Second, ECL$^{-s,-h}$ is the worst. Among the three ablation models, ECL$^{-s,-h}$ removes the contrastive learning on simple and hard negative samples together from the standard ECL, which rollbacks to the original BART model. Without the support of contrastive learning framework, its performance is hurt greatly. 

The above experimental results imply that each key component in ECL, including hard and simple negative samples, and contrastive learning framework, is necessary for the outstanding performance of ECL.

\subsection{Comparison of Focus Identification Rate}
\label{sec:focusidentification}

Among the baselines introduced in Section \ref{sec:baselines}, QFCL is the state-of-the-art model. Both QFCL and our proposed ECL model attempt to generate hard negative samples according to question focus. However, their detailed strategies are different. Specifically, QFCL assumes the
overlap phrases between CHQ and FAQ as question focuses, while our ECL method assumes medical entities in
FAQs as question focuses. Whether they can accurately recognize focus of medical questions will directly affect the generation of hard negative samples. 

To investigate the influence of the two strategies in QFCL and ECL, we define the focus identification rate (FIR), which measures how many questions can be identified focuses, i.e., $FIR = \frac{|D^{focus}|}{|D|}$, where $|D|$ denotes the number of questions in the dataset $D$,  and $|D^{focus}|$ refers to the number of questions whose focuses can be identified in $D$. 

\begin{figure}[htbp]
\centering
\includegraphics[width=0.5\textwidth]{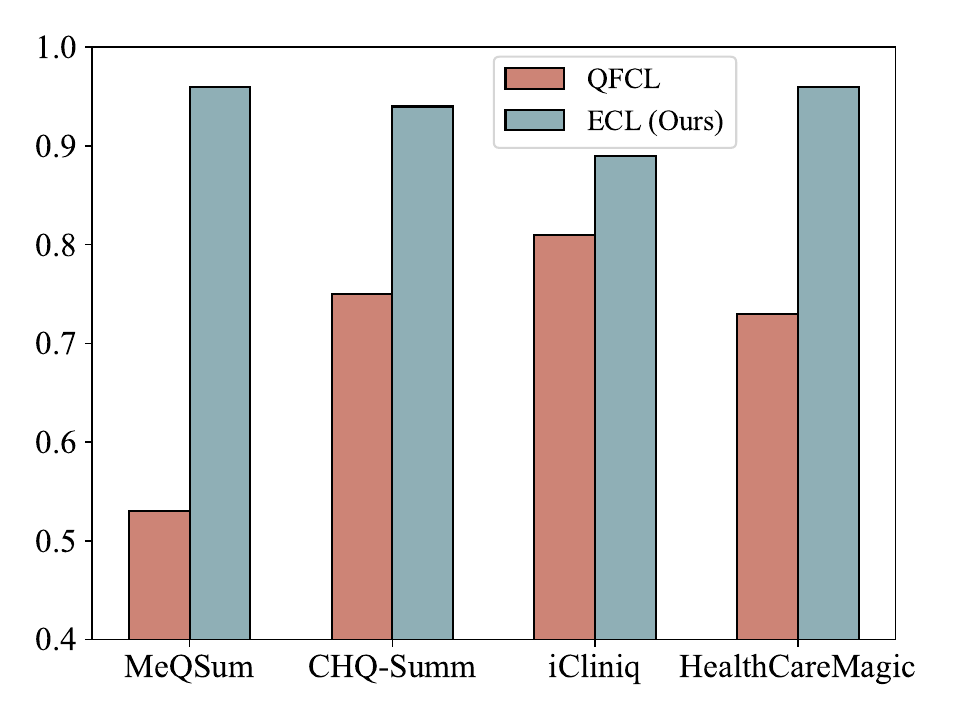}

\caption{Comparison of Focus Identification Rate (FIR). \blue{The x-axis represents the dataset, and the y-axis represents the FIR}}
\label{ner}
\end{figure}

We utilize the focus identification strategy in QFCL and ECL respectively, to extract question focuses in the training datasets, and report the FIR metric in Fig.~\ref{ner}. According to the figure, on four datasets, the focus identification rates of QFCL are 0.53, 0.75, 0.81, and 0.73 respectively, while those of ECL are 0.96, 0.94, 0.89, and 0.96 respectively. The significant gap between QFCL and ECL demonstrates that our ECL method is more effective, which can benefit the construction of hard negative samples.

\subsection{Comparison of Medical Entity Consistency}
\label{sec:medicalentityconsistency}

To investigate whether our ECL model can identify medical entities and generate summaries containing right medical entities, we define the evaluation metric of medical entity consistency, as below:

\begin{equation}
Consistency = \frac{{\sum\limits_{i = 1}^N {\left| {E(S_i^{ref}) \cap E(S_i^{gen})} \right|} }}{{\sum\limits_{i = 1}^N {\left| {E(S_i^{ref})} \right|} }},
\end{equation}
where $N$ is the size of the test dataset, ${E(S_i^{ref})}$ and ${E(S_i^{gen})}$ represent the medical named entities in the reference summaries and generated summaries, respectively. The experimental results are shown in Table ~\ref{capture}.

According to the table, comparing with the popular BART and QFCL, our ECL model can achieve more higher consistency of medical entities on most datasets. Specifically, ECL gains the best performance on the MeQSum, CHQ-Summ and HealthCareMagic datasets, which achieves 0.75\%, 0.75\% and 1.01\% improvements over the suboptimal model, respectively. On the iCliniq dataset, ECL is slightly inferior to the others, but it is still comparable. Overall, the experimental results demonstrate that our ECL model is more effective than the others on the metric of entity consistency, which can pay more attention to medical entities and generate summaries with more right medical entities. 

\begin{table*}[htbp]
    \caption{Comparison of Medical Entity Consistency.}
    \centering
    \begin{tabular}{l|c|c|c|c}
        \hline
        Model & MeQSum & CHQ-Summ & iCliniq & HealthCareMagic \\ \hline
        \hline
        BART & 45.05\%  &\underline{33.42\%} &\textbf{35.94\%} & 30.72\% \\ \hline 
        QFCL & \underline{45.54\%}  &31.4\% &\underline{35.86\%} & \underline{31.20\%} \\ \hline
        ECL(ours) & \textbf{46.29\%}(+0.75\%) & \textbf{34.17\%}(+0.75\%) & 35.72\%(-0.22\%)  & \textbf{32.21\%}(+1.01\%) \\ \hline
    \multicolumn{5}{l}{\makecell[l]{*The best performance is boldfaced, and the suboptimal one is underlined.}}
    \end{tabular}
    \label{capture}
\end{table*}

\subsection{Analysis on Contrastive Learning with Negative Samples}
\label{sec:CLwithnegativesamples}
\begin{figure*}[htbp]
\centering
\includegraphics[width=0.85\textwidth]{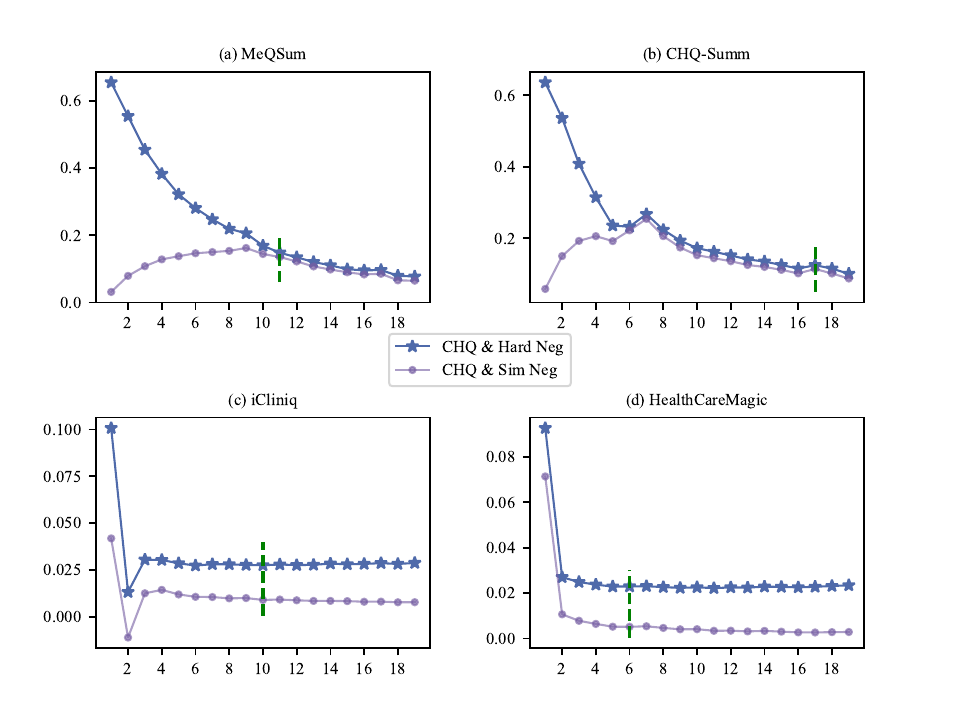}
\caption{The procedure of similarity optimization during ECL training. \blue{The x-axis represents epoches of model training, and the y-axis represents similarities between CHQ and hard/simple negative samples.} The dotted line in green indicates that the model performs best on the validation set at the corresponding epoch.}
\label{fig:sim}
\end{figure*} 

According to Section \ref{sec:scl}
and \ref{sec:hcl}, besides the cross-entropy loss ${{\mathcal{L}}_{ce}}$, our ECL model employs the losses of contrastive learning on simple and hard negative samples, i.e., ${{\mathcal{L}}_{scl}}$ and ${{\mathcal{L}}_{hcl}}$, to jointly optimize the parameters. The purpose of ${{\mathcal{L}}_{scl}}$ and ${{\mathcal{L}}_{hcl}}$ is to train the model to decrease the semantic similarity between CHQs and negative samples, including sample and hard ones. Intuitively, if the model are trained well, the similarity between CHQs and negative samples should decreased gradually with the increasement of training epochs. 

To investigate whether our ECL model keeps consistency with the former intuition, we utilize cosine similarity to evaluate semantic similarity between CHQs and negative samples, and draw the procedure of similarity optimization during ECL training, as shown in Fig.~\ref{fig:sim}. According to the figure, we have the following observations. 

First, for the similarity of CHQs and hard negative samples, it decreases gradually with the increasement of training epochs on four datasets. This is consistent with the intuition. This demonstrates that ${{\mathcal{L}}_{hcl}}$ is workable, which is able to train the model to decrease the similarity of CHQs and hard negative samples. 

Second, for the similarity of CHQs and sample negative samples, it also decreases gradually on the iCliniq and HealthCareMagic datasets, which meets with the intuition. However, it first shows an unexpected increasement, and then shows the expected decrease on the MeQSum and CHQ-Summ datasets. The reason for the unexpected increasement may be that there exists parameter competition between ${{\mathcal{L}}_{scl}}$ and ${{\mathcal{L}}_{hcl}}$ when optimizing model parameters. They influence each other, but have not reached a better balance point. Although there exists unexpected increasement in the early training on two datasets, the overall similarities still decrease with the increasement of training epochs, which also demonstrates the effectiveness of ${{\mathcal{L}}_{scl}}$.

Third, compared with the MeQSum and CHQ-Summ datasets, ECL works more stably on the iCliniq and HealthCareMagic datasets. This may be because that the MeQSum and CHQ-Summ datasets only contain 1,000 and 1,500 CHQs respectively, which are very small, compared with the other two datasets. It is reasonable that ECL can be trained more stably on the two larger datasets. 

\subsection{Case Study}
\label{sec:casestudy}
\begin{table*}[htbp]
    \caption{Examples of generated question summaries by BART, QFCL and our ECL model. Medical entities in medical questions are \blue{highlighted with red color}.}
    \centering
    \begin{tabular}{|c|l|}
        \hline
         \multicolumn{2}{|l|}{\textbf{Case 1}} \\
         \hline\hline
         CHQ & \makecell[l]{SUBJECT: Arthritis\\MESSAGE: Is the medication  \textcolor{red}{Trimethoprim/sulfamethozle} useful in \textcolor{red}{arthritis}?    Thank you} \\
         \hline
         FAQ & Is \textcolor{red}{Trimethoprim/sulfamethazole} indicated for \textcolor{red}{arthritis}? \\
         \hline\hline
         BART & What are the side effects of \textcolor{red}{Trimethoprim/sulfamethozle}? \\ 
         \hline
         QFCL & Is \textcolor{red}{Trimethoprim/sulfamethozle} effective in \textcolor{red}{arthritis}? \\
         \hline
         ECL(ours) &  Is \textcolor{red}{Trimethoprim/sulfamethozle} an effective treatment for \textcolor{red}{arthritis}? \\
         \hline\hline
         \multicolumn{2}{|l|}{\textbf{Case 2}} \\ \hline\hline 
         CHQ & \makecell[l]{SUBJECT: dressings\\MESSAGE: my dad had some skin cancer removed 3 wks ago and a \textcolor{red}{skin graft} was performed. The do-\\ner site is healing well, but the graft site is still very sore and having \textcolor{red}{dressings} changed every 2/3 days,\\ dad isnt convinced that the wet gauze \textcolor{red}{dressings} are any good. Which type of \textcolor{red}{dressings} should be being\\ used?} \\ \hline
        FAQ & What types of \textcolor{red}{dressings} are used for \textcolor{red}{skin graft} sites? \\ \hline \hline
         BART & What are the treatments for \textcolor{red}{skin grafts}? \\  \hline
         QFCL & What are the treatments for \textcolor{red}{skin grafts}? \\
         \hline
         ECL(ours) & What type of \textcolor{red}{dressings} should be applied after \textcolor{red}{skin graft}? \\ \hline\hline
         \multicolumn{2}{|l|}{\textbf{Case 3}} \\
         \hline\hline
         CHQ & \makecell[l]{SUBJECT: Gout and Blood pressure\\MESSAGE: I have had many gout attacks since i have been 30 years old now 70. I take \textcolor{red}{allopurinol} and\\ bloodpressure meds . Before i took \textcolor{red}{allopurinol} i never had high blood pressure. Also i have developed b-\\asal cell skin cancer i have heard \textcolor{red}{allopurinol} will cause that . Reduces acid in your system?}\\
         \hline
         FAQ & What are the \textcolor{red}{side effects} of \textcolor{red}{allopurinol}? \\
         \hline
         \hline
         BART & What are the treatments for gout and blood pressure? \\ 
         \hline
         QFCL & What are the treatments for gout and blood pressure? \\
         \hline
         ECL(ours) &  What are the \textcolor{red}{side effects} of \textcolor{red}{allopurinol}? \\
         \hline
         
    \end{tabular}
    \label{case}
\end{table*}

To clearly show the summaries generated by different methods, we list three samples of question summaries to compare our ECL model with BART and QFCL, as shown in Table ~\ref{case}. 

In Case 1, BART generates the medical entity \textit{Trimethoprim/ sulfamethozle}, but misses the medical entity \textit{arthritis}. In contrast, both QFCL and ECL can capture all medical entities successfully.

In Case 2, BART and QFCL generate the medical entity \textit{skin grafts}, but miss the medical entity \textit{dressings}. In contrast, ECL can capture all medical entities.

In Case 3, BART and QFCL generate the summaries with wrong medical entities, i.e., \textit{gout} and \textit{blood pressure}, which miss the right entities., i.e., \textit{allopurinol} and \textit{side effects}. In contrast, ECL can generate an ideal summary with all necessary entities. 

The cases mentioned above demonstrate that our ECL model is more powerful than the popular state-of-the-art models, i.e., BART and QFCL, which can pay more attention to medical entities in CHQs and generate better summaries with medical entities. The reason for the superiority of ECL lies in that it is built on a framework of entity-driven contrastive learning, which equips ECL with the ability of recognizing and generating medical entities in CHQs and their summaries.

\subsection{Discussion}
According to the experiments in Section \ref{sec:experimentalresult}, we can observe that the methods based on pretrained models, i.e., BART, perform well on medical question summarization. Among these methods, our ECL model can achieve the best performance, thanks to our proposed hard negative sample generation mechanism based on medical entities. Specifically, the hard negative samples generated by replacing medical entities have very similar encoding representations to the source medical question, but differ greatly in terms of semantics. We utilize the golden reference summary of the source medical question as the positive sample, and employ the contrastive learning framework to minimize the distance between the encoding representation of the source medical question and its golden reference summary, while maximizing the distance between the encoding representation of the source medical question and the hard negative samples. In this way, we can succeed to improve the quality of the encoder's encoding and implicitly train the encoder to pay more attention to the medical entities of the medical question, i.e., the focus. Moreover, we randomly select a large number of simple negative samples for contrastive learning. Comparing with hard negative samples, although the quality of simple ones is low, using a large number of simple negative samples can still improve the model's performance to some extent, as shown in the ablation experiments in Section \ref{sec:ablationstudy}. 

Furthermore, according to the experiments on focus identification rate in Section \ref{sec:focusidentification}, we can clearly observe that using medical entities as the focus of medical questions is more effective than selecting overlapping words. According to the experimental results on medical entity consistency in Section \ref{sec:medicalentityconsistency}, it is obvious that the ECL model has better consistency on most datasets, which also indicates that our method indeed implicitly trains the model to pay more attention to the medical entities in medical questions. Lastly, the experiments in Section \ref{sec:CLwithnegativesamples} analyze the role of contrastive learning with negative samples in the training process of the ECL model, and the results show that contrastive learning can succeed to maximize the distance between CHQs and negative samples, which can achieve more stable performance on larger datasets. 

\section{Conclusions}
Medical question summarization (MQS) is critical yet challenging as MQS is often embedded in medical answering systems to facilitate the retrieval of correct answers. Existing work is confused by the obvious distinctions on health question descriptions from patients and doctors, and is troubled by the absence of enough reliable evaluation datasets. This paper proposes a framework for medical question summarization with entity-driven contrastive learning (ECL). ECL devises an easy and effective mechanism to capture question focus with medical entity, and then utilizes them to generate hard negative samples more accurately. Besides, the data leakage problem of iCliniq and HealthCareMagic datasets are corrected, which become reliable and can evaluate models fairly. Exhaustive experiments demonstrate that ECL can achieve significant improvement over the state-of-the-art baselines. This may provide a new perspective for enhance the contrastive learning in MQS task. Our future work is to investigate more effective way to generate more harder negative samples with medical knowledge graph, and to apply ECL on MQS task in other languages, such as Chinese.

\section*{Acknowledgment}
This work was supported by the National Natural Science Foundation of China (Grant No.62376130), Shandong Provincial Natural Science Foundation (Grant No.ZR2022MF243), Program of New Twenty Policies for Universities of Jinan (Grant No.202333008), Program of Innovation Improvement of Shandong (Grant No.2023TSGC0182).

\bibliographystyle{ACM-Reference-Format}
\bibliography{main}
\end{document}